\documentclass[sigconf]{acmart}

\AtBeginDocument{%
  \providecommand\BibTeX{{%
    \normalfont B\kern-0.5em{\scshape i\kern-0.25em b}\kern-0.8em\TeX}}}

\setcopyright{acmcopyright}
\copyrightyear{2018}
\acmYear{2018}
\acmDOI{XXXXXXX.XXXXXXX}

\acmConference[Conference acronym 'XX]{Make sure to enter the correct
  conference title from your rights confirmation emai}{June 03--05,
  2018}{Woodstock, NY}
\acmPrice{15.00}
\acmISBN{978-1-4503-XXXX-X/18/06}

\usepackage{colortbl}
\usepackage{stfloats}
\usepackage{subfigure}
\usepackage{algorithm}
\usepackage{algorithmic}

\begin{document}

\title{CMMD: Cross-Metric Multi-Dimensional Root Cause Analysis}
\author{Shifu Yan$^{1*}$, Caihua Shan$^2$, Wenyi Yang$^2$, Bixiong Xu$^2$, Dongsheng Li$^2$, Lili Qiu$^2$, Jie Tong$^2$, Qi Zhang$^2$ }\thanks{*The work was done during the author's  internship with Microsoft.}
\affiliation{%
\institution{$^1$East China University of Science and Technology, $^2$Microsoft \\
 	shifuyan677@hotmail.com, \{caihuashan, wenya, bix, dongsheng.li, liliqiu, jietong, qizhang\}@microsoft.com}
 \country{}
 }
 
\newcommand{\stitle}[1]{\vspace{1ex}\noindent\textup{\textbf{#1.}}}
\renewcommand{\shortauthors}{Yan, et al.}

\begin{abstract}


In large-scale online services, crucial metrics, a.k.a., key performance indicators (KPIs), are monitored periodically to check their running statuses. 
Generally, KPIs are aggregated along multiple dimensions and derived by complex calculations among fundamental metrics from the raw data. Once abnormal KPI values are observed, root cause analysis (RCA) can be applied to identify the reasons for anomalies, so that we can troubleshoot quickly. 
Recently, several automatic RCA techniques were proposed to localize the related dimensions (or a combination of dimensions) to explain the anomalies. 
However, their analyses are limited to the data on the abnormal metric and ignore the data of other metrics which may be also related to the anomalies, leading to imprecise or even incorrect root causes.
To this end, we propose a cross-metric multi-dimensional root cause analysis method, named CMMD, which consists of two key components: 1) relationship modeling, which utilizes graph neural network (GNN) to model the unknown complex calculation among metrics and aggregation function among dimensions from historical data; 2) root cause localization, which adopts the genetic algorithm to efficiently and effectively dive into the raw data and localize the abnormal dimension(s) once the KPI anomalies are detected. 
Experiments on synthetic datasets, public datasets and online production environment demonstrate the superiority of our proposed CMMD method compared with baselines. Currently, CMMD is running as an online service in  Microsoft Azure. 

\end{abstract}



\keywords{Root cause analysis, graph neural network, genetic algorithm}

\maketitle

\section{Introduction}



\begin{figure}[h!]
 \centering
 \includegraphics[width=\linewidth]{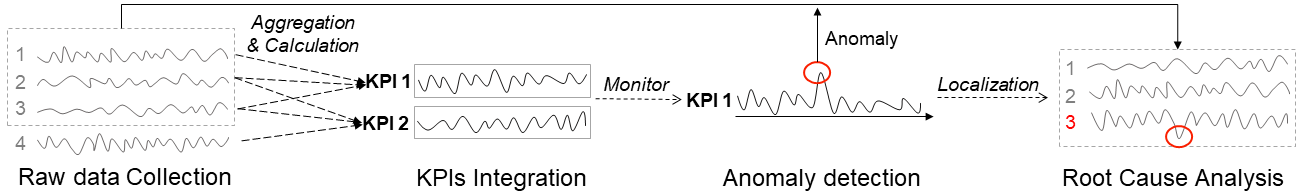}
 \caption{General pipeline of KPI monitoring.}
 \label{fig:pipeline}  
\end{figure}

In the digital era, many companies and organizations monitor the status of their products, services, and business through data intelligence. A general pipeline is shown in Fig.~\ref{fig:pipeline}: 1) collect the huge volume of data streams; 2) integrate the raw data into KPI metrics related to the performance and reliability issues; 3) monitor KPI metrics periodically and detect the anomalies intelligently; 4) diagnose root causes to prevent anomalies becoming customer-impacting incidents. In this paper, we focus on the last but the essential part, root cause analysis (RCA).
Formally, given the multi-dimensional streaming data as input, RCA localizes which dimension(s) of the data as the reason to the anomaly detected in KPIs.

\begin{figure*}[h!]
 \centering
 \includegraphics[width=0.9\linewidth]{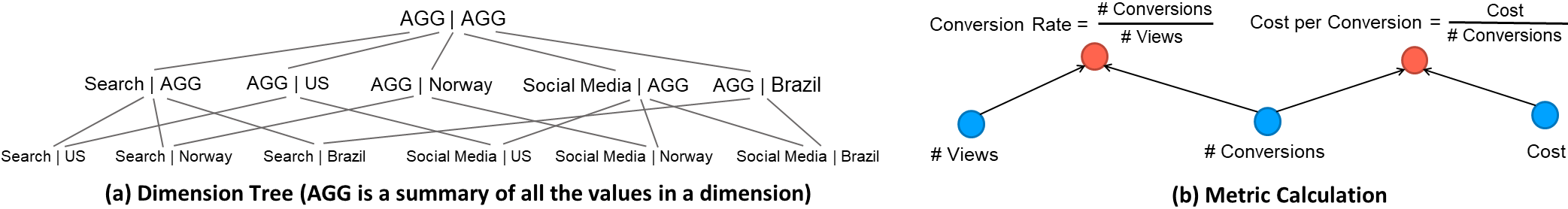}
 \caption{Dimensions and metrics of a business system: an example.}
 \label{fig:dimension_and_metric}  
\end{figure*}

There are two main challenges because the raw data contains many multi-dimensional metrics, while the monitored KPI metric is highly aggregated and often derived.
For a certain metric, there could be multiple dimensions, e.g., region and channel. There exist numerous values in each dimension, such as Region=\{US, UK,...\}.
Given $L$ dimensions and $V$ values in each dimension, we have $V^L$ dimension value combinations (e.g., Region=US $\cap$ Channel=Search). 
It is too resource-consuming to monitor each of the combinations independently when $L$ and $V$ are large. 
Thus, the best practice is to aggregate the value along the dimension. Fig.~\ref{fig:dimension_and_metric}(a) shows a dimension tree, and ``AGG'' indicates we conduct an aggregation function (e.g., SUM or MEAN) on all values in that dimension. As long as the root node is monitored, we can check the status of all the dimensions. Overall, the first challenge for RCA is how to accurately and quickly find a set of dimension value combinations (leaf nodes) to resolve the anomaly detected in the root node.

Besides, the monitored KPI metric often requires the complex calculation of the raw data. Some applications include the total error rate to check whether a large-scale software is running normally, profitability index to decide the performance of a company product, and overall equipment effectiveness to evaluate how well a manufacturing operation is utilized. Take a business system as a detailed instance: the system monitors the conversion rate of advertisements as its KPI to measure the success of advertisements. The raw streaming data contain the fundamental metrics (e.g., $\#$views, $\#$conversions), and the derived KPI metrics (e.g., conversion rate = $\frac{\#\text{conversions}}{\#\text{views}}$). It is obvious that the deviations of derived KPI metrics are caused by the changes of fundamental metrics. Thus, the second challenge is how to explain the anomaly occurred in the derived metric by considering all the metric data, even if we do not know the exact calculation among metrics.

\begin{table*} \small
  \caption{A Snapshot of business metrics at timestamp $t$. We show the real value (forecast value in parentheses) for each metric in each possible dimension. 
  The monitored metric Conversion Rate is abnormal because of a 27$\%$ decrease (colored in red). Previous methods found two root causes colored in yellow. CMMD believed the real root cause is Search|US colored in green.}
  \label{tab:metrics}
  \begin{tabular}{cc|ccc|cc}
    \toprule
    \multicolumn{2}{c}{\textbf{Dimensions}} & \multicolumn{3}{c}{\textbf{Fundamental Metrics}}& \multicolumn{2}{c}{\textbf{Derived Metrics}}\\
    Channel &  Region & $\#$Views  & $\#$Conversions & Cost & Conversion Rate & Cost per Conversion\\
    \midrule
    Search & US & 51949(57328) & \cellcolor[rgb]{0,0.9,0.2}\textbf{14651(25741)} & 219765(249067) & 0.28(0.45) & 15(17)\\
    Search & Norway & 3152(2627) & 783(1228) & 13311(12528) & \cellcolor[rgb]{0.8,0.8,0}\textbf{0.25(0.47)} & 17(16)\\
    Search & Brazil & 3125(2981) & 341(980) & 6820(7502) & \cellcolor[rgb]{0.8,0.8,0}\textbf{0.11(0.33)} & 20(22)\\
    Search & Others & 64351(59721) & 19321(25931) & 618272(579630) & 0.30(0.43) & 32(30)\\ 
    Social Media & US & 43949(39312) & 21525(24057) & 344400(322875) & 0.49(0.59) & 16(15)\\
    Social Media & Norway & 20453(18327) & 8731(9068) & 139696(148427) & 0.43(0.50) & 16(17)\\
    Social Media & Brazil & 1957(1512) & 1023(1001) & 17391(16368) & 0.52(0.66) & 17(16)\\
    Social Media & Others & 70384(60413) & 32253(35912) & 903084(838578) & 0.46(0.59) & 28(26)\\
    AGG & US & 95898(96640) & 36176(49798) & 564165(614992) & 0.38(0.50) & 16(17)\\
    AGG & Norway & 23605(20954) & 9514(10296) & 153007(161738) & 0.40(0.49) & 16(17)\\
    AGG & Brazil & 5082(4493) & 1364(1981) & 24211(25916) & 0.27(0.44) & 18(19)\\
    AGG & Others & 134735(120134) & 51574(61843) & 1521356(1598794) & 0.38(0.50) & 29(31)\\
    Search & AGG & 122577(122657) & 35096(53880) & 858168(877400) & 0.28(0.43) & 24(25)\\
    Social Media & AGG & 136743(119564) & 63532(70038) & 1404571(1397704) & 0.46(0.59) & 22(22)\\
    \midrule
    AGG & AGG & 259320(242221) & 98628(123918) & 2262739(2268444) & \cellcolor[rgb]{1,0,0}\textbf{0.38(0.52)} & 22(23)\\
  \bottomrule
\end{tabular}
\end{table*}

Recently, several works were proposed to automatically identify root causes by regarding each root cause as the combination of values from different dimensions, which cannot address the above challenges.
For instance, iDice~\cite{lin2016idice} and HotSpot~\cite{sun2018hotspot} only works when the monitored metric is fundamental. Adtributor~\cite{bhagwan2014adtributor} and Squeeze~\cite{li2019generic} considered the derived metrics. But Adtributor~\cite{bhagwan2014adtributor} assumes the root cause is only in one dimension. Squeeze~\cite{li2019generic} supports multiple dimension value combinations, but it needs to know the number of root causes as a prior. 

Most importantly, all the existing methods focused on the single-metric instead of cross-metric scenario, 
i.e., they only consider the data related to the abnormal metric and ignore the data of other metrics which are also related to the anomaly. 
We illustrate it with the following example. Tab.~\ref{tab:metrics} shows a snapshot of a business system. We sum up all the values in a dimension as a tree based on Fig.~\ref{fig:dimension_and_metric}(a), and calculate two derived metrics Conversion Rate and Cost per Conversion based on Fig.~\ref{fig:dimension_and_metric}(b).
To see whether the system is running normally, we monitor two KPIs, Conversion Rate of AGG|AGG and Cost per Conversion of AGG|AGG. 
At timestamp t, an anomaly is detected in Conversion Rate of AGG|AGG due to a 27\% decrease (0.52 $\rightarrow$ 0.38). Traditional methods only focus on the changes of conversion rate (the data in the column of conversion rate), so they regard the root cause as Search|Norway and Search|Brazil since 47\% and 67\% decline are detected in their conversion rate (0.47 $\rightarrow$ 0.25, 0.33 $\rightarrow$ 0.11). 
However, from the absolute value of \#Views and \#Conversions, we find that the decrease of \#Conversions of Search|US contributes most to the anomaly.
If we recover \#Conversions of Search|US (14651 $\rightarrow$ 25741), \#Conversions of AGG|AGG becomes 109718, and then Conversion Rate of AGG|AGG is changed from 0.38 to 0.42. If we recover Search|Norway and Search|Brazil, Conversion Rate of AGG|AGG is almost not changed. Search|US has a higher recovery ratio. In summary, this example shows that it is hard to find the accurate root cause when fundamental metrics are not considered.

To this end, we propose a \textbf{c}ross-\textbf{m}etric \textbf{m}ulti-\textbf{d}imensional root cause analysis approach, named CMMD, which can identify precise root causes with fewer assumptions. 
Firstly, we do not assume that we know the exact calculation among metrics and aggregation function among dimensions. We automatically model the relationship among metrics of different dimensions by graph neural network from historical data. 
Secondly, there is no assumption of the number or range of root causes. 
We adopt the genetic algorithm and design a novel fitness score based on the trained GNN to compute the probability of a candidate to be a real root cause. The candidate could be any possible dimension value combinations of all the metrics related to the anomaly. Besides, we also utilize the attention weights in the trained GNN to speed up. 
Thirdly, CMMD could do fine-grained analyses for the raw data. Since we aggregate all the dimensions as a tree, we can monitor the status of any node at any level, not only the root node. Further, after we obtain the root causes searched by the genetic algorithm, we backtrack and aggregate these root causes in the dimension tree to make the result succinct and fine-grained.
Finally, CMMD is evaluated by three types of dataset. For online production datasets where anomalies are detected in the derived metric while root causes are in the fundamental metric, we achieve at least 9\% improvement compared with baselines. For public datasets where anomalies and root causes are both in the same metric, a competitive performance is also exhibited. The major contributions are summarized as follows:

\begin{itemize}
\item To our best knowledge, CMMD is the first RCA approach for the cross-metric scenario and can analyze the root causes for both fundamental and derived metrics.
\item We model the unknown relationship among metrics and dimensions based on graph neural networks. 
\item We utilize the genetic algorithm with designed fitness score to search out the precise root cause.
\item We demonstrate the industrial practice in Microsoft Azure.

\end{itemize}

\section{Related works}

In this section, we  introduce existing approaches for root cause analysis and multivariate time-series tasks, and  summarize the differences between these approaches and CMMD.
\begin{table*} \scriptsize 
  \caption{Notations and corresponding examples used in this paper.}
  \label{tab:notations}
  \begin{tabular}{c|c|c}
    \toprule
    Notation & Definition & Example\\
    \midrule
    $a_{ij}$ & a value of each dimension & US, Norway for Region; Social Media, Search for Channel
    \\
    $\mathbb{T}$ &  dimension tree   & An aggregated tree among dimensions\\
    a node in $\mathbb{T}$ &  a dimension value combination  & Social Media|US, Search|Norway, AGG|US\\
    root cause &  a set of dimension value combinations & \{Search|US, Social Media|AGG\}\\
    $M^F$ / $M^D$ & fundamental metric / derived metric & $\#$Views, $\#$Conversions, Cost / Conversion Rate, Cost per Conversion\\
    $v_{i,m}$ / $f_{i,m}$  & real/expected values of the metric $m$ in the node $i$ & In Tab.~\ref{tab:metrics}, $v = 341 $ and
     $f = 980$ of the node Search|Brazil for the metric \#Conversions\\
  \bottomrule
\end{tabular}
\end{table*}

\subsection{Root Cause Analysis} 


The goal of root cause analysis in multi-dimensional data is to localize a combination of dimension values to explain the anomaly that happened in the monitored metric.  Adtributor~\cite{bhagwan2014adtributor} is firstly proposed under the assumption that the root cause lies in only one dimension. Based on the Jensen-Shannon divergence between real and expected values of abnormal metrics, Adtributor selected the dimension value with the highest divergence as the root cause. Adtributor can be applied recursively, named RAdtributor, to find several dimension values to be the root cause. 
iDice~\cite{lin2016idice} removed the assumption and considered the root cause is a combination of dimension values, i.e., the anomaly occurs with the coaction of many dimensions. Thus, iDice utilized a tree-based method to evaluate the dimension value combinations by Fisher distance between real and expected values of abnormal metrics. At the same time, the pruning strategy is also adopted to filter the most effective dimension combination. ImAPTr~\cite{rong2020locating} used a similar tree-based method to analyze root causes for declining success rate and replaced Fisher distance with Jensen-Shannon divergence.
HotSpot~\cite{sun2018hotspot} defined \textit{ripple effect} to propagate the anomaly into different dimensions and proposed \textit{potential score} to evaluate a set of dimension value combinations by replacing their real values with corresponding expected values and calculating the difference with or without the replacement. HotSpot also adopted the Monte Carlo Tree Search and a hierarchical pruning strategy to improve search efficiency. 
According to HotSpot, Squeeze~\cite{li2019generic} proposed \textit{generalized ripple effect} and \textit{generalized potential score} for both fundamental and derived metric. Differently, Squeeze first grouped potential abnormal dimension value combinations into clusters based on generalized ripple effect, and then searched for the root causes based on generalized potential score within those clusters. 
To improve the clustering performance in Squeeze, AutoRoot~\cite{jing2021autoroot} used the kernel density estimation to cluster adaptively and proposed a new method to calculate generalized potential score for insignificant deviation magnitude. However, Squeeze and AutoRoot need to assume the number of root causes in a cluster as a prior. 
Given the dependency among dimensions, HALO~\cite{zhang2021halo} formed a hierarchy structure of dimensions instead of the dimension tree based on conditional entropy and searched root causes by walking in this structure. MID~\cite{gu2020efficient} improved the efficiency of iDice by an evolution-based search framework. HALO and MID are specially designed for the cloud system.

iDice and HotSpot work when the monitored metric is fundamental, while Adtributor, Squeeze and AutoRoot support the analysis of derived metrics. However, all the existing methods are focused on the single-metric situation. They only consider the multi-dimensional data of the abnormal metric and ignore the data of other metrics which are also related to the anomalies. CMMD takes all the multi-dimensional data among various metrics into consideration and thus  
achieves more precise root cause analysis.  

\subsection{Multivariate Time-series Tasks}

\stitle{Forecasting} Time-series forecasting aims to better understand the data by predicting future values in the time series. Traditional univariate time-series methods, such as auto regression (AR)~\cite{gunnemann2014robust}, moving average (MA)~\cite{nakano2017generalized} and auto regressive integrated moving average (ARIMA)~\cite{box1970distribution} predict the data behavior based on the historical data in a linear way. Considering nonlinearity, neural networks including CNN~\cite{munir2018deepant} and LSTM~\cite{park2018multimodal} have been successfully applied. Recently, some GNN-based methods~\cite{wu2020connecting,cao2021spectral} were proposed to capture both intra-series and inter-series correlations for multivariate time-series forecasting, which obtained outstanding performance.

Different from multivariate time-series data, CMMD is focused on multi-dimensional streaming data. If there are millions or billions of dimension value combinations, it is impossible to train a multivariate model to learn the relationship in practice. Thus, based on the observation that multi-dimensional data is conducted by the same aggregation function and calculation at each timestamp, we simplify the problem and propose a single-layer GNN model. 
Note that this observation is universal in many industrial applications. 

\stitle{Anomaly Detection} Many multivariate time-series anomaly detection approaches~\cite{park2018multimodal,deng2021graph} have been studied recently. However, we regard the module of anomaly detection as a black box in the paper. We consider the situation where the anomaly is already detected. 

\section{Problem formulation}

Here we formulate the problem for cross-metric multi-dimensional root cause analysis. Tab.~\ref{tab:notations} lists the notions and examples. In the raw streaming data, there are multiple dimensions and each dimension contains numerous values. We denote $a_{ij}$ as the $j$-th value in the $i$-th dimension. Besides, the raw data also contain two kinds of metrics, fundamental metrics $M^F=\left\{m^F_1,m^F_2,...,m^F_p \right\}$ and derived metrics $M^D=\left\{m^D_1,m^D_2,...,m^D_q \right\}$. At each timestamp, we have the real numerical value of each $a_{ij}$ for every metric $m$ as the input.

Dimension values are often aggregated as a dimension tree $\mathbb{T}$, as shown in Fig.~\ref{fig:dimension_and_metric}.
AGG represents a summary of all the values in a dimension. 
Each node in $\mathbb{T}$ is a combination with specific values in the dimension, i.e., a vector $\left[a_{11},a_{23},...,a_{ij},...\right]$. 
For non-leaf nodes, $a_{ij}$ in some positions could be AGG, which represents that the $i$-th dimension is already aggregated. For the root node, all the positions in the vector are AGG that means all the dimensions are aggregated.

After that, the most representative status, the derived KPI metric of the root node is monitored periodically. When the anomaly is detected by Anomaly Detection module, the main objective of Root Cause Analysis is to analyze which set of dimension value combinations is the root cause of the anomaly. In other words, we need to identify a set of nodes in $\mathbb{T}$ as the root cause, such as 
$\left\{\left[a_{11},a_{21},... \right], \left[a_{14}, a_{27},...\right],... \right\}$.

\begin{figure}[t!]
  \centering
  \includegraphics[width=\linewidth]{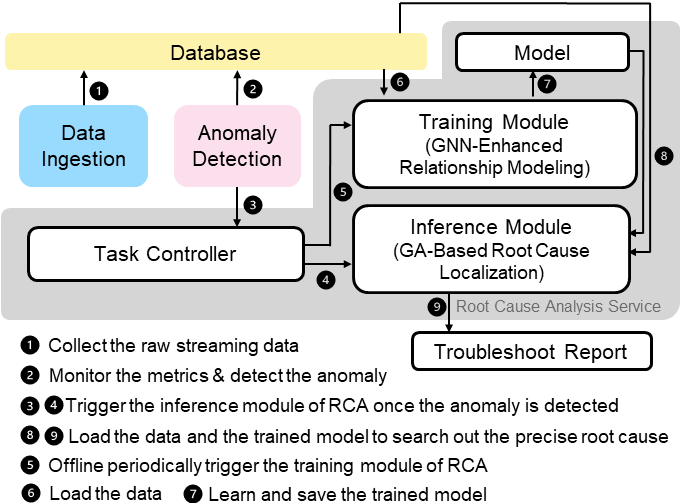}
  \caption{Framework of the proposed method. The main algorithm is described by the components colored in grey.}
  \label{fig:overview}  
\end{figure}

\section{Methodology}
In the section, we first present the framework of CMMD, and then introduce the two important modules in CMMD, training and inference modules. Finally, we describe two extra functions for fine-grained analysis and speed-up.   
\subsection{Framework}
The overall framework of CMMD is presented in Fig.~\ref{fig:overview}. In the stream processing, we collect and store the raw data in the database. Anomaly detection is used to monitor KPI metrics integrated from the streaming data. Once it detects the anomaly, it sends the signal to the root cause analysis service. RCA is batch-processing where tasks are managed in the task controller. There are two kinds of tasks: 1) we offline periodically trigger the training module to train/update the model for the relationship between the monitored metric and the fundamental metrics obtained from the raw streaming data; 2) once there exists the anomaly signal, we trigger the inference module to search out the precise root cause and output a troubleshoot report. 


In detail, we construct a dimension tree by aggregating dimensions in the offline training module, and a graph attention network~\cite{velivckovic2017graph} is used to learn the tree-based relationship. In the online inference module, a bottom-up search strategy is adopted based on the genetic algorithm~\cite{whitley1994genetic}. We design a fitness function based on the graph attention model to evaluate the probability of the candidate dimension value combination to be a root cause.

\begin{figure*}[h]
  \centering
  \includegraphics[width=1.0\textwidth]{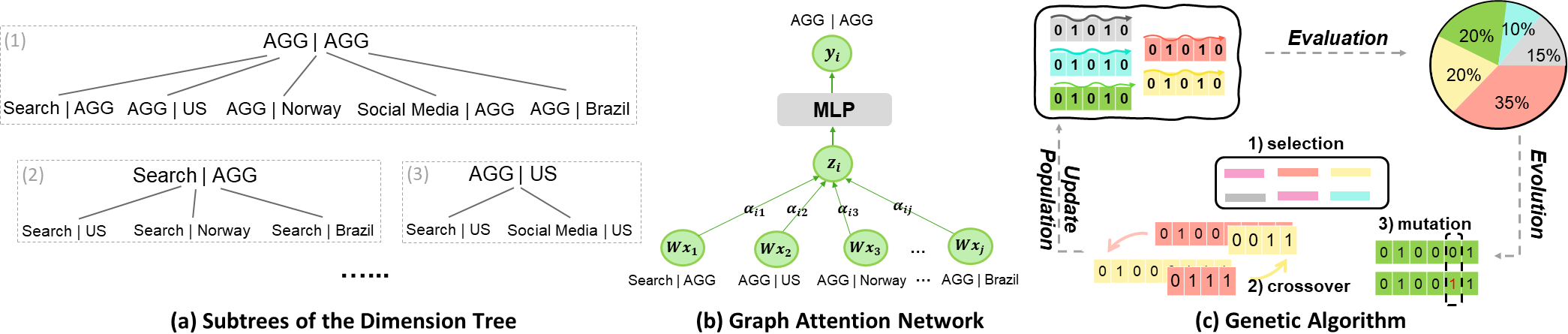}
  \caption{The key components  of CMMD.}
  \label{fig:modules}  
\end{figure*}

\subsection{GNN-Enhanced Relationship Modeling} \label{sec:relationship_modeling}

We often monitor the root node in the dimension tree for derived metrics. However, we need to localize the root cause in the raw streaming data, i.e., the leaf nodes for fundamental metrics. Thus, the first step is to model the complex relationship. In this section, we introduce how to utilize the graph attention network (GAT) to learn the relationship from historical data automatically.

Although we do not know the detailed calculation used to get the value of derived metrics from fundamental metrics and the exact aggregation function to aggregate the value along the dimension, we find the calculation and aggregation are the same in any two-layer subtree. As shown in Fig.~\ref{fig:modules}(a), we extract several two-layer subtrees from the dimension tree in Fig.~\ref{fig:dimension_and_metric}. The data processing in each subtree is the same: at each timestamp, the raw data are collected in child nodes for fundamental metrics. Then these data are aggregated along the dimensions to generate the value of the parent node. Meanwhile, the value of fundamental metrics is calculated to obtain the value of derived metrics in each node.
Since the number of child nodes is uncertain, graph neural network is suitable to learn the pattern in the subtree.

Formally, we define $v_{t,i,m}$ as the values of node $i$ for each metric $m$ at timestamp $t$. Suppose that we have $L$ dimensions, $P$ fundamental metrics and $Q$ derived metrics. For each node $i$, the value of fundamental metrics is the input $x_i = [v_{t,i,m^F_1}, v_{t,i,m^F_2},..., v_{t,i,m^F_P}]$, and the value of derived metrics is the output $y_i = [v_{t,i,m^D_1}, v_{t,i,m^D_2},..., v_{t,i,m^D_Q}]$. 
Since the relationship is the same at each timestamp, we omit the subscript $t$. For each subtree, we utilize the \textbf{single-layer graph attention} $\mathcal{G}$ to model the relationship between the child nodes and the parent node: $y_i = \mathcal{G}(\{x_j | \text{node $i$ is the parent of node $j$}\})$. If we conduct $L$ graph convolutions, we iteratively learn the relationship between the leaf nodes and the root node in the dimension tree, i.e., $y_\text{root} = \mathcal{G}^L(\{x_j |\text{node $j$ is the leaf node}\})$.

\stitle{Single-layer graph attention} 
According to the tree structure, we specify the neighbors $\mathcal{N}$ as the child nodes of each parent node. As shown in Fig.~\ref{fig:modules}(b), the representation $z_i$ of node $i$ after message passing is calculated as follows:
\begin{equation} \small
    z_i = \sigma\left(\sum_{j \in \mathcal{N}_i}\alpha_{ij} \boldsymbol{W} x_j \right),
\end{equation}
where $\boldsymbol{W}$ and $\sigma$ are the parameters and activation function in GAT. Please note that we add the depth of node $j$ into the input feature $x_j$ to enrich the information. The attention coefficients $\alpha_{ij}$ between node $i$ and node $j$ can be computed as follows:
\begin{equation} \small
    \alpha_{ij} = \frac{exp\left(LeakyReLU\left(\boldsymbol{\theta}^T\left[\boldsymbol{W}x_i\|\boldsymbol{W}x_j \right] \right) \right)}{\sum_{k\in\mathcal{N}_i}exp\left(LeakyReLU\left(\boldsymbol{\theta}^T\left[\boldsymbol{W}x_i\|\boldsymbol{W}x_k \right] \right) \right)},
\end{equation}
where $\boldsymbol{\theta}$ is the learnable parameter of a single-layer feedforward neural network indicating the attention mechanism and $\|$ means concatenation of vectors. We also implement the multi-head attention to learn more useful information:
\begin{equation} \small
    z_i = \|^K_{k=1}\sigma\left(\sum_{j\in\mathcal{N}_i}\alpha^k_{ij}\boldsymbol{W}^k x_j \right).
\end{equation}

\stitle{Objective function}
After aggregating information from child nodes and obtain the representations $z_i$, we finally apply a multi-layer perceptron to fit the function between the representations and the real value of derived metrics $\hat{y} \in \mathbb{R}^q$. Here we use the Mean Squared Error (MSE) to guide the GAT training:
\begin{equation} \small
    \textit{L}_{MSE} = \sum_{t=1..T}\sum_{i=1..N'}||\hat{y}_{t,i} - MLP\left(z_{t,i} \right)||_2^2,
\end{equation}
where $T$ is the total number of timestamps and $N'$ is the number of non-leaf nodes in the dimension tree.

\subsection{GA-Based Root Cause Localization} \label{sec:evolutionary_search}

Based on the above model, we are able to get the effect of fundamental metrics in the leaf nodes on a specific monitored metric. Therefore, we can analyze the causes for the abnormal metrics once detected. However, it is time-consuming because of a large scale of dimension value combinations in practice. In this way, tree-based search methods are popular in root cause analysis. 
If we utilize the top-down search strategy, there may exist many assumptions and pruning thresholds, and some root causes will be missing. Thus, we introduce a heuristic search framework based on the genetic algorithm for this combinatorial optimization problem and the bottom-up search strategy is applied with few assumptions about the root cause.

\begin{figure}[ht!]
  \centering
  \includegraphics[width=\linewidth]{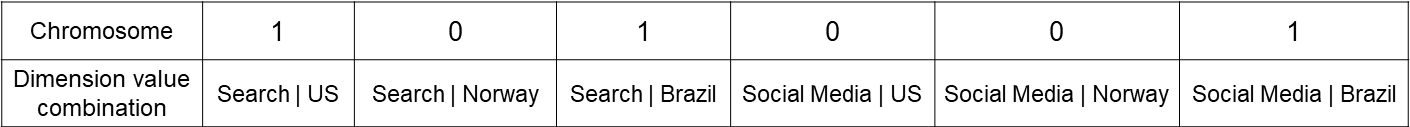}
  \caption{Chromosome Encoding.}
  \label{fig:encode}  
\end{figure}
Anomaly detection in a metric is based on the deviation between the real value $v$ and the predicted value $f$. Such prediction can be implemented by most of the time-series algorithms (e.g., AR, MA, ARIMA). Considering the efficiency and accuracy in root cause analysis, AR is adopted to obtain the forecast value of each node in this work.
When an anomaly is detected, we conduct the GA-based search which is shown in Fig.~\ref{fig:modules}(c).

\stitle{Encoding} 
First of all, we encode root cause candidates, i.e., all the leaf nodes in the dimension tree, into a vector (chromosome). As shown in Fig.~\ref{fig:encode}, the chromosome is a binary bit string, which can be represented as:
\begin{equation}
    s = [s_1, \cdots, s_i, \cdots, s_n] \quad  s_i \in \left\{0,1 \right\}, 1 \leq i \leq n.
\end{equation}
Each bit $s_i$ indicates the status of a dimension value combination, such as Search|US. $s_i=1$ means that this dimension value combination is chosen as the root cause and $s_i=0$ means that it is excluded. Obviously, the number of root causes can be more than one, which means that the abnormality is due to the coaction of multiple abnormal dimension value combinations. Note that we do not assume the number of root causes, and it is decided in the search process automatically.


\stitle{Evaluation}
Based on the encoding mechanism, we randomly sample multiple chromosomes as an initial population. The next step is to compute the probability of each chromosome to be an accurate root cause set. We design the fitness score, whose motivation is to evaluate whether the abnormality still exists if we 'troubleshoot' the root cause set indicated by the chromosome.  

We denote the deviation as $\Delta_{\text{root},m} = |v_{\text{root},m} - f_{\text{root},m}|$ where $v_{\text{root},m}$ and $f_{\text{root},m}$ are the real value and forecast value of the root node for the abnormal metric $m$. Since the chromosome $s$ indicates abnormal dimension value combinations, we replace the values of abnormal dimension value combinations with the predicted values, maintain the values of normal dimension value combination, and obtain the new input $\boldsymbol{X}(s)$. Next, we could obtain the new output value $\mathcal{G}^L(\boldsymbol{X}(s))$ based on the graph model $\mathcal{G}^L$ learned in Sec.~\ref{sec:relationship_modeling}.
To evaluate whether eliminating the abnormal dimension value combinations indicated by $s$ is enough to troubleshoot the target metric, we define the fitness score as:
\begin{equation} \small
    \text{Fitness score} = \frac{|\mathcal{G}^L\left(\boldsymbol{X}(s) \right) - f_{\text{root},m}|}{|v_{\text{root},m} - f_{\text{root},m}|} + \beta ||s||_1.
    \label{eq:fs}
\end{equation}
Obviously, chromosome with smaller score is considered to be the root cause set with higher probability. Besides, we need to consider the succinctness of the root causes to avoid selecting all of them. $\beta$ is a trade-off parameter between the deviation and succinctness.

\stitle{Evolution}
The core of the genetic algorithm is the iterative evaluation and evolution which is made up of three behaviors, selection, crossover and mutation.
 
Given a population with the fitness score of each chromosome, the \textbf{selection} operator aims to choose chromosomes for the next iteration based on the roulette and a chromosome with a smaller fitness score is more likely to be selected obviously. 
After selection, we perform \textbf{crossover} and \textbf{mutation} to enrich chromosome diversity. The crossover operation is controlled by the cross rate $t_c$. For a pair of chromosomes, if the probability we randomly generated exceeds $t_c$, we swap part of these two chromosomes at a random bit. $t_c$ determines the frequency of new chromosomes. A small $t_c$ leads to the slow generation of new chromosomes and affects the diversity. Meanwhile, a large $t_c$ results in the unstable heredity of chromosomes and makes GA similar to random algorithm. Therefore, we set $t_c=0.5$. The mutation operation is controlled by another mutation rate $t_f$. Similarly for each chromosome, if the probability exceeds $t_f$, then we randomly pick a bit in this chromosome and reverse 0 and 1. Since mutation is a small probability event and the influence of mutation should be much less than crossover, $t_f$ is usually recommended to be less than 0.2 and we set $t_f=0.1$. We summarized the whole search process in Alg.~\ref{algo:GA} in Appendix.

\subsection{Extra Functions}
\subsubsection{Backtrack for Fine-grained Analysis}
In the previous sections, we find out the root causes based on the optimal chromosome $s^*$ with the smallest fitness score. Furthermore, we could obtain the fine-grained result by aggregating these chosen leaf nodes into their parent nodes.
For example, if the leaf nodes Social Media|US and Search|US are both chosen as the root cause, the result can be compacted as their parent node, AGG|US. To achieve this goal, we backtrack in the dimension tree. 

We summarized the whole process in Alg.~\ref{algo:backtrack} in Appendix. Assume that we have a set of chosen leaf nodes $\mathcal{S}$ and a backtrack threshold $t_\gamma$ as input. Then, we conduct a bottom-up aggregation layer by layer. For a parent node, if the percentage of its child nodes in $\mathcal{S}$ is more than $t_\gamma$, we remove these child nodes from $\mathcal{S}$ and add the parent node into $\mathcal{S}$. $t_\gamma$ is a user-defined which determines the succinctness of results and large $t_\gamma$ will make root causes include more leaf nodes. $t_\gamma\geq0.5$ is usually preferred and we set $t_\gamma=0.6$ in the experiments. We stop when there is no parent node added in a layer. Finally, the new root cause set $\mathcal{S}^*$ is succinct. 

\subsubsection{Filtering for Speed-up}
The efficiency of root causes analysis depends largely on the number of dimension value combinations, i.e., the chromosome length. Given the fact that not all dimension value combinations are abnormal when the anomaly is detected, we filter out the normal dimension value combinations to obtain the candidates for the genetic algorithm. Considering the deviation between real value $v_{i,m}$ and expected value $f_{i,m}$ of the leaf node $i$ of the abnormal metric $m$, and the importance of leaf node $i$ to the root node, we design the following score to measure whether it can be a candidate:
\begin{equation} \small
    {\text{Filtering score}}_i = \frac{|v_{i,m}-f_{i,m}|}{|v_{i,m}|} * \frac{\alpha_{i,\text{root}}}{\sum_j \alpha_{j,\text{root}}}.
    \label{eq:filter}
\end{equation}
The importance to the root node $\alpha_{i,\text{root}}$ is computed iteratively by the average $k$-head attention score $\frac{1}{K}\sum_k\alpha^k_{ij}$ in the graph model. If $\text{filtering score}_i$ is smaller than the filtering threshold $t_\delta$, we filter the node $i$. 


\section{Experiment}
In the section, we conduct extensive experiments to verify both the effectiveness and efficiency of the proposed CMMD on multiple datasets. Due to the space limitation, we move the implementation detail (including hyper-parameter setting of CMMD and baselines) and parameter study to Appendix.
\subsection{Experimental setup}

\subsubsection*{\textbf{Datasets}}
To comprehensively assess the effectiveness of our method, we conduct experiments on three types of situations, production datasets (obtained from online services at Microsoft), public datasets (released by the previous work) and synthetic datasets.

\stitle{Production datasets $\mathcal{D}_1$}
Here we have 5 production cases. The first 3 cases came from a business analysis scenario. Conversion rate, cost per conversion, and cost per view are three important metrics to track the marketing effort and analyze the results. They are generated by three fundamental metrics, the number of conversions, the number of views and cost.
The calculations are 
conversion rate = $\frac{\text{\#conversions}}{\text{\#views}}$, 
cost per conversion = $\frac{\text{cost}}{\text{\#conversions}}$, 
and cost per view = 
$\frac{\text{cost}}{\text{\#views}}$.
We collected 3-month data with 91 timestamps, consisting of two dimensions (Region and Channel). There are 249 countries or regions in Region and 3 values in Channel. Except the undefined combinations, there are 503 leaf dimension value combinations, and 
270 aggregated dimension value combinations in the tree. 

The remaining 2 cases are from an alerting service. We gathered 1-year metrics with 300 timestamps. There is one dimension notification channel which has 5 different values. Alert open rate and click-through rate are monitored to bring to light the quality and the engagement of the alerts, which is calculated based on the number of alerts, the number of opened alerts and the number of click-through alerts. The calculations are 
alert open rate = 
$\frac{\text{\#opened alerts}}{\text{\#alerts}}$, 
and click-through rate =
$\frac{\text{\#click-through alerts}}{\text{\#alerts}}$. 


\stitle{Public datasets $\mathcal{D}_2$}
The public dataset is released by Squeeze~\cite{li2019generic}. $\mathcal{D}_2$ is a semi-synthetic dataset constructed based on the real-world dataset from an Internet Company. Here we select its derived sub-dataset as $\mathcal{D}_2$. The derived metric is the success rate in transactions which is calculated by successful transaction number and total transaction number. There are 100 timestamps in each case with 4 dimensions and 21600 leaf dimension value combinations. In this way, anomalies are injected as (\textit{p},\textit{q}) where \textit{p}, \textit{q} mean the number of abnormal dimensions and the number of root causes respectively. We set \textit{p} = 1,2,3 and \textit{q} = 1,2,3 to test 9 cases respectively.

\stitle{Synthetic datasets $\mathcal{D}_3$}
Apart from the production and public datasets, we also conduct experiments on a synthetic dataset $\mathcal{D}_3$. We have two fundamental metrics $b$ and $c$, and two derived metrics $a=g(c)$ and $d = f(a,b)$.
For evaluation on more complex relationships among metrics, we set function $f$ as

1) $\frac{a}{b}$, 2) $a \times b$, 3) $\frac{\log a}{\log b}$, 4) $a \times e^b$ and 5) $\frac{\log (a+1)}{\log (b+1)}$. The function $g$ is randomly selected from $\left\{c, \sin(c), e^c, c^2, \sqrt{c} \right\}$.
In each case, 200 timestamps are simulated. The detailed simulation process is described in Appendix. 

\subsubsection*{\textbf{Evaluation metrics}}
In the following experiments, F1-score is used to evaluate the performance in cross-metric multiple dimensional root cause analysis. Since current methods only focus on the dimensions, we only compare the result whether the root causes are the accurate set of dimension value combinations. The calculation of F1-score is mathematically described as follows.
\begin{equation} \small
    \text{F1 score} = \frac{2*\text{Precision}*\text{Recall}}{\text{Precision} + \text{Recall}},
\end{equation}
where $\text{Precision (P)} = \frac{\text{TP}}{\text{TP} + \text{FP}}$ and $\text{Recall (R)} = \frac{\text{TP}}{\text{TP} + \text{FN}}$. TP (true positive) is the number of dimension value combinations both reported by the algorithm and the ground truth. FP (false positive) is the number of dimension value combinations reported by algorithm but not in the ground truth. FN (false negative) means the number of root causes in ground truth but not reported by the algorithm. 

The ground truth of $\mathcal{D}_2$ is provided in \cite{li2019generic}. In other two datasets, since we know the exact calculation among metrics and aggregation function among dimensions,
the ground truth is determined by the recovery ratio of specific dimension value combinations. Please see the detailed description in Appendix. 

\subsubsection*{\textbf{Comparative methods}}
We compare the proposed CMMD with four baselines, Adtributor~\cite{bhagwan2014adtributor}, HotSpot~\cite{sun2018hotspot}, Squeeze~\cite{li2019generic} and AutoRoot~\cite{jing2021autoroot}. The detailed information for baselines is introduced in Appendix. 


\subsection{Overall Performance}
Firstly, we conduct experiments on the production datasets $\mathcal{D}_1$ and present the results of the compared methods in Tab.~\ref{tab:expprod}. In the results, CMMD outperforms other methods in these five cases and achieves nearly 9\% higher than the second-best performance on average F1 score. In these methods, Squeeze obtains the highest Recall, especially in the case of cost per conversion. HotSpot, Squeeze and AutoRoot exhibit similar F1 scores with competitive Recall in industrial practice, but the Precision of the detected root causes needs to be further improved especially for Squeeze. These methods do not consider the cross metrics which will result in multiple false positives since many abnormal dimension value combination does not contribute to the monitored metrics. Although Adtributor provides a better Precision than HotSpot, Squeeze and AutoRoot, the Recall of root causes cannot meet the requirements of practical application which means some root causes are undetected by Adtributor. Overall, the proposed CMMD provides the best performance and demonstrates that fundamental metrics can improve the accuracy of root cause analysis when monitoring the derived metrics.

\begin{table*}[t] \footnotesize
  \caption{Experimental results on production datasets.}
  \label{tab:expprod}
  \begin{tabular}{c|ccc|ccc|ccc|ccc|ccc}
    \toprule
      & \multicolumn{3}{c}{Adtributor} & \multicolumn{3}{c}{HotSpot} & \multicolumn{3}{c}{Squeeze} & \multicolumn{3}{c}{AutoRoot} & \multicolumn{3}{c}{CMMD} \\
    Datasets & P & R & F1 & P & R & F1 & P & R & F1 & P & R & F1 & P & R & F1 \\
    \midrule
    Conversion rate & 0.7679 & 0.5939 & 0.6698 & 0.7036 & 0.8591 & 0.7736 & 0.6136 & 0.7459 & 0.6733 & 0.6005 & 0.7182 & 0.6541 & 0.7339 & 0.9144 & \textbf{0.8143} \\
    Cost per conversion & 0.7202 & 0.6205 & 0.6667 & 0.6572 & 0.7733 & 0.7105 & 0.6844 & 0.8902 & 0.7739 & 0.6693 & 0.8019 & 0.7296 & 0.8054 & 0.8496 & \textbf{0.8269} \\
    Cost per view & 1.0000 & 0.5000 & 0.6667 & 1.0000 & 0.6154 & 0.7619 & 1.0000 & 0.8077 & 0.8936 & 0.9524 & 0.7693 & 0.8511 & 1.0000 & 0.8462 & \textbf{0.9167} \\
    Alert open rate & 0.4400 & 0.3200 & 0.3721 & 0.4800 & 0.9600 & 0.6400 & 0.4386 & 1.0000 & 0.6098 & 0.4310 & 1.0000 & 0.6024 & 0.8333 & 0.8000 & \textbf{0.8164} \\
    Click-through rate & 1.0000 & 0.3143 & 0.4783 & 1.0000 & 0.7714 & 0.8525 & 1.0000 & 0.8571 & 0.9231 & 1.0000 & 0.8286 & 0.9063 & 1.0000 & 0.8857 & \textbf{0.9394} \\
    \midrule
    Average & 0.7856 & 0.4697 & 0.5707 & 0.7682 & 0.7958 & 0.7477 & 0.7473 & 0.8602 & 0.7747 & 0.7306 & 0.8236 & 0.7487 & 0.8745 & 0.8592 & \textbf{0.8627} \\
    \bottomrule
  \end{tabular}
\end{table*}

\begin{table*}[h] \footnotesize
  \caption{Experimental results on public real-world datasets.}
  \label{tab:exppub}
  \begin{tabular}{c|ccc|ccc|ccc|ccc|ccc}
    \toprule
        & \multicolumn{3}{c}{Adtributor} & \multicolumn{3}{c}{HotSpot} & \multicolumn{3}{c}{Squeeze} & \multicolumn{3}{c}{AutoRoot} & \multicolumn{3}{c}{CMMD} \\ 
        Root cause type & P & R & F1 & P & R & F1 & P & R & F1 & P & R & F1 & P & R & F1 \\
    \midrule
        (1,1) & 0.8000 & 0.9500 & 0.8686 & 0.7882 & 1.0000 & 0.8815 & 0.9200 & 0.9200 & 0.9200 & 0.9000 & 0.9000 & 0.9000 & 0.9000 & 0.9500 & \textbf{0.9243} \\
        (1,2) & 0.9855 & 0.7083 & 0.8240 & 0.5394 & 0.7500 & 0.6275 & 0.9540 & 	0.8300 & 0.8877 & 0.9253 & 0.8050 & 0.8610 & 0.9242 & 0.9150 & \textbf{0.9196} \\
        (1,3) & 0.9171 & 0.6700 & 0.7743 & 0.3322 & 0.6575 & 0.4414 & 0.9336 & 0.7226 & 0.8146 & 0.9115 & 0.7055 & 0.7954 & 0.8793 & 0.8500 & \textbf{0.8644} \\
        (2,1) & 0.4826 & 0.8384 & 0.6125 & 0.6899 & 0.7200 & 0.7042 & 0.8700 & 0.8200 & 0.8443 & 0.8700 & 0.8700 & 0.8700 & 0.8519 & 0.9200 & \textbf{0.8846} \\
        (2,2) & 0.5472 & 0.7323 & 0.6264 & 0.4462 & 0.5200 & 0.4803 & 0.8955 & 0.9000 & \textbf{0.8977} & 0.8955 & 0.9000 & \textbf{0.8977} & 0.8974 & 0.8750 & 0.8861 \\
        (2,3) & 0.6735 & 0.5556 & 0.6089 & 0.4181 & 0.4000 & 0.4088 & 0.8930 & 0.8930 & \textbf{0.8930} & 0.8930 & 0.8930 & \textbf{0.8930} & 0.8970 & 0.8700 & 0.8833 \\
        (3,1) & 0.2908 & 0.8200 & 0.4294 & 0.4502 & 0.5300 & 0.4869 & 0.8700 & 0.8900 & 0.8799 & 0.8911 & 0.9000 & \textbf{0.8955} & 0.8762 & 0.8500 & 0.8629 \\
        (3,2) & 0.4255 & 0.6061 & 0.5000 & 0.3500 & 0.4100 & 0.3776 & 0.9250 & 0.8900 & 0.9059 & 0.9200 & 0.9200 & 0.9200 & 0.9333 & 0.9200 & \textbf{0.9266} \\
        (3,3) & 0.4580 & 0.4511 & 0.4545 & 0.3550 & 0.3533 & 0.3541 & 0.8896 & 0.8867 & \textbf{0.8881} & 0.8896 & 0.8867 & \textbf{0.8881} & 0.8616 & 0.8800 & 0.8707 \\
    \midrule
        Average & 0.6200 & 0.7035 & 0.6332 & 0.4855 & 0.5934 & 0.5291 & 0.9056 & 0.8614 & 0.8812 & 0.8996 & 0.8645 & 0.8801 & 0.8912 & 0.8922 & \textbf{0.8914} \\
    \bottomrule
  \end{tabular}
\end{table*}

Next, we evaluate these five methods on the public datasets $\mathcal{D}_2$. The results in Tab.~\ref{tab:exppub} suggest that CMMD can still exhibit better F1 scores on average. It is worth noting that the anomalies are injected directly into the derived metrics rather than the fundamental metrics and this mechanism leads to a single-metric scenario. 
Therefore, HotSpot, Squeeze and AutoRoot for the single-metric task can also exhibit competitive performance. Similarly, Adtributor is not effective in Recall. With the increasing number of dimensions in a root cause \textit{p} and the increasing number of root causes \textit{q}, the performance of all methods gradually deteriorates. 
The clustering algorithms in Squeeze and AutoRoot can help determine the number of root causes and thus they have better results compared with Adtributor and HotSpot. Although CMMD is
not good as AutoRoot in some cases, it still obtains a competitive performance. 

\begin{figure*}[h]
  \centering
  \subfigure[Precision]{
  \begin{minipage}[b]{0.22\textwidth}
  \label{fig:syn1}
  \includegraphics[width=\linewidth]{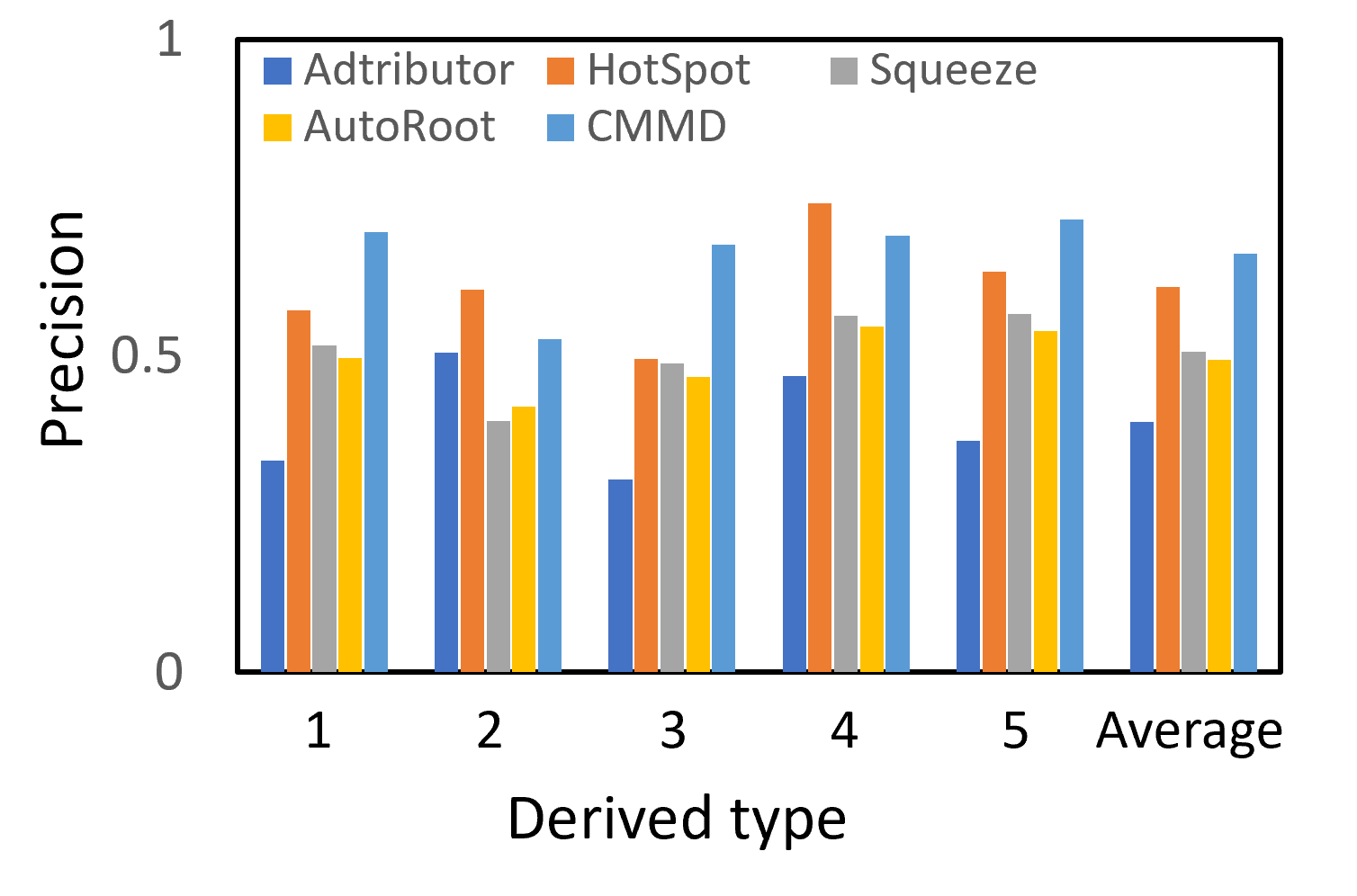}
  \end{minipage}
  }
  \subfigure[Recall]{
  \begin{minipage}[b]{0.22\textwidth}
  \label{fig:syn2}
  \includegraphics[width=\linewidth]{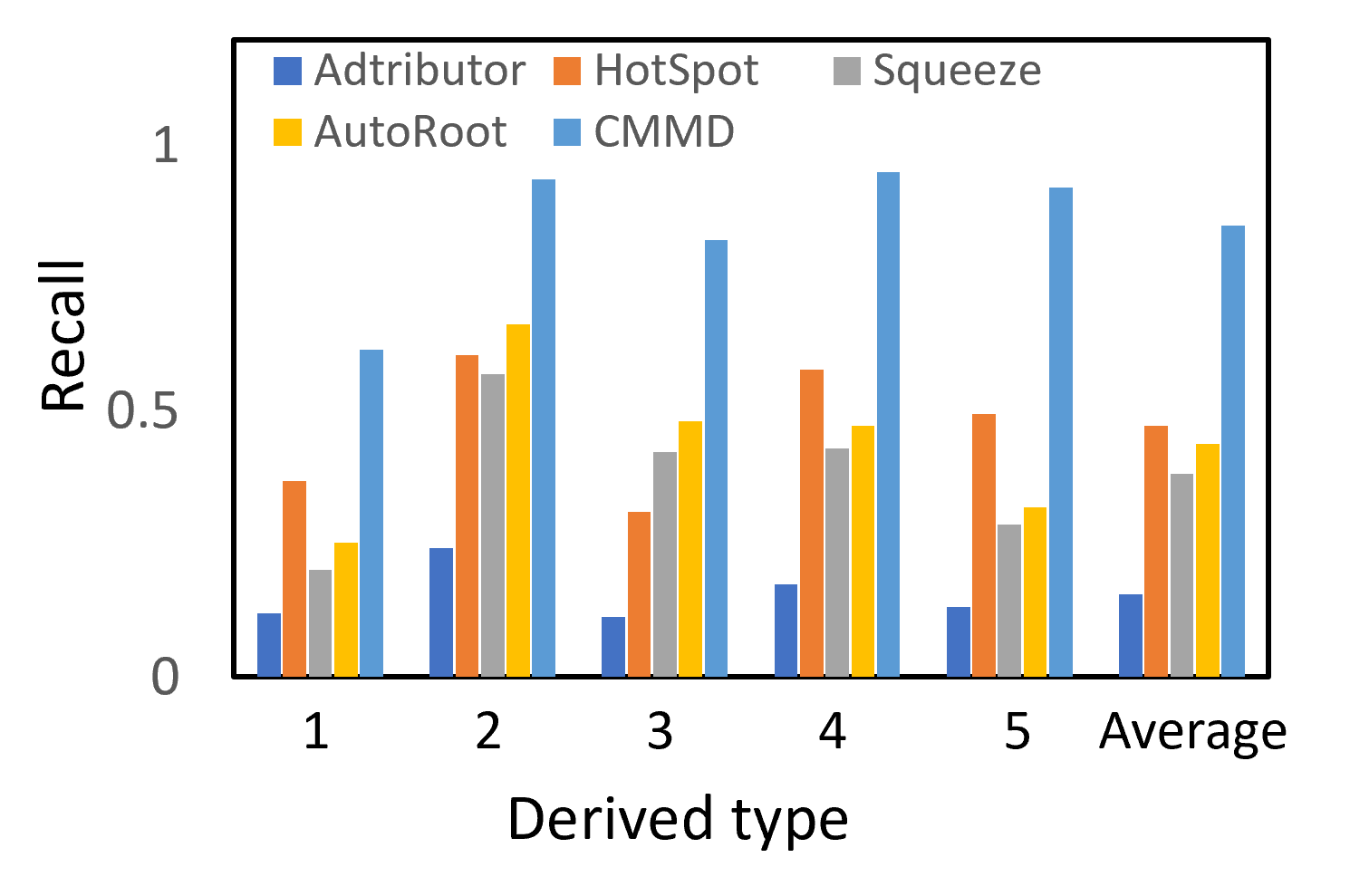}
  \end{minipage}
  }
  \subfigure[F1 score]{
  \begin{minipage}[b]{0.22\textwidth}
  \label{fig:syn3}
  \includegraphics[width=\linewidth]{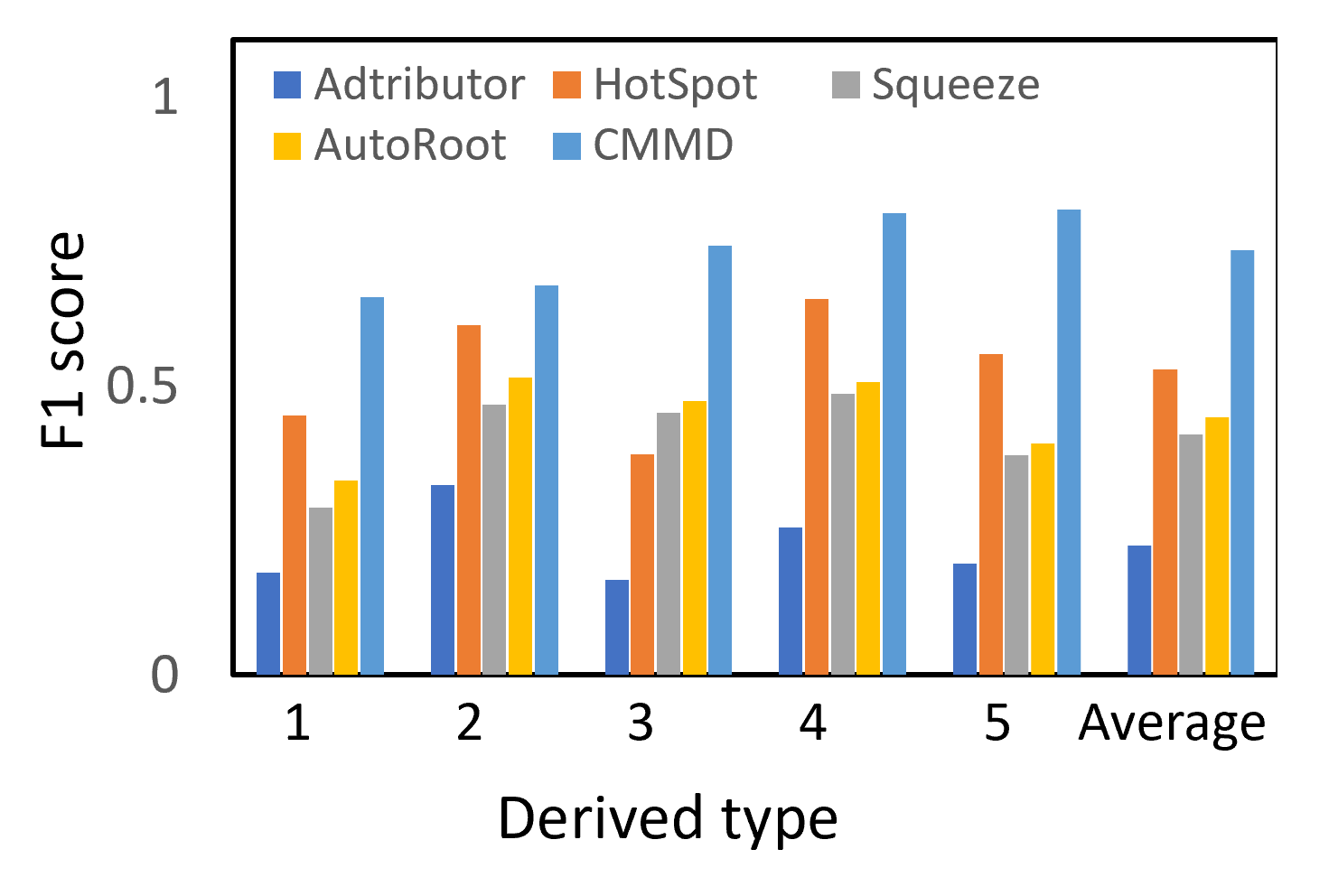}
  \end{minipage}
  }
  \subfigure[Inference time]{
  \begin{minipage}[b]{0.22\textwidth}
  \label{fig:time}
  \includegraphics[width=\linewidth]{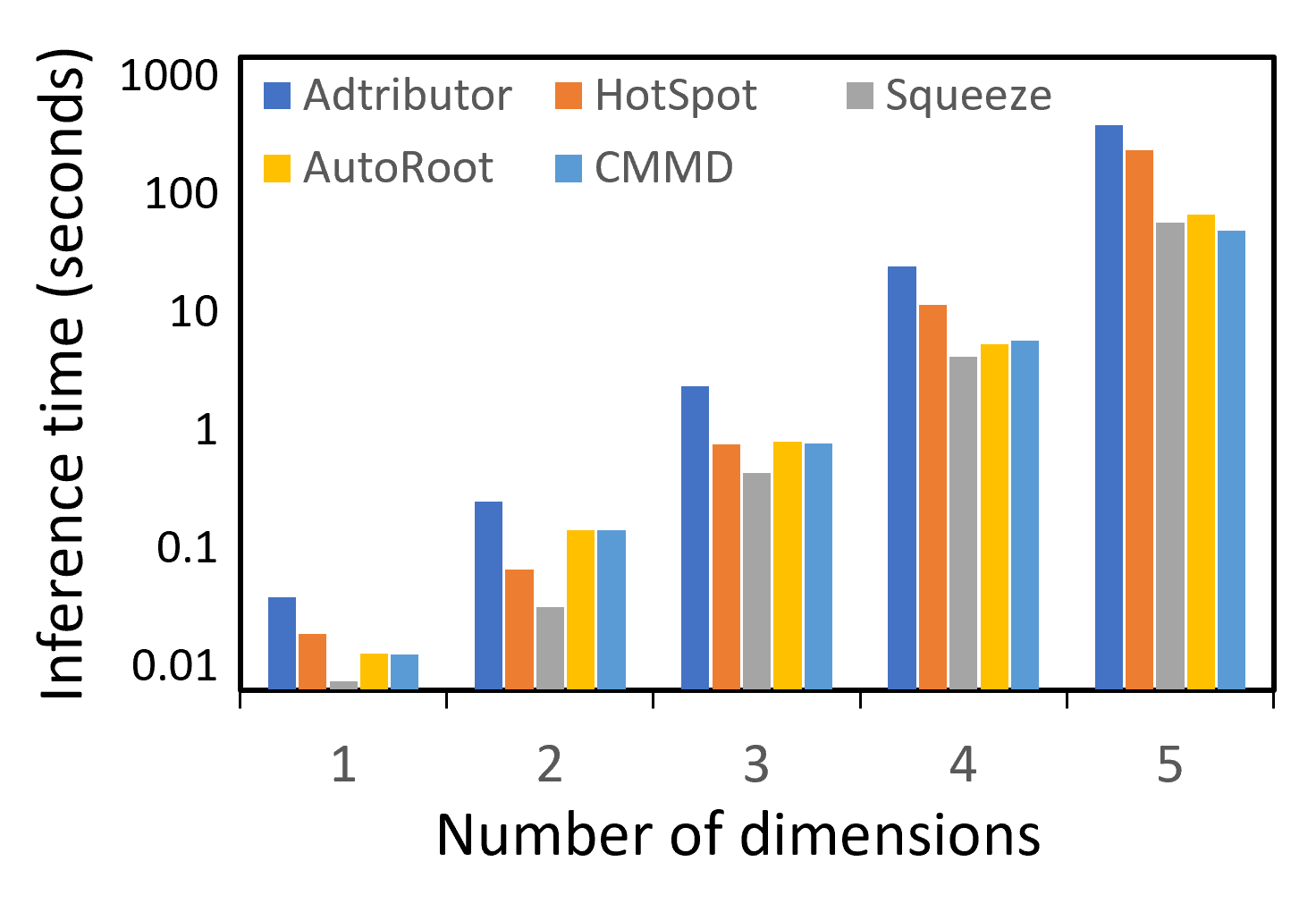}
  \end{minipage}
  }
  \caption{Experimental results on synthetic datasets.}
  \label{fig:syn}  
\end{figure*}

Finally, we provide the Precision, Recall and F1 score in Fig.~\ref{fig:syn}(a-c) for the five synthetic datasets. Overall, CMMD can still achieve better F1 scores and is flexible for different calculations between fundamental and derived metrics. In these datasets, Adtributor exhibits a better Precision but a bad Recall since it tends to consider fewer root causes. The performance of Squeeze and AutoRoot is more balanced in Precision and Recall. HotSpot presents a closer performance to our method with higher Precision in a few cases. The good performance on synthetic datasets demonstrates the generalizability of our proposed CMMD.

Besides, the efficiency of root cause analysis is also important for industrial practice. We also conduct experiments on synthetic datasets with an increasing number of dimensions and set each dimension consisting of ten dimension values. Therefore, we evaluate these methods with up to $10^5$ leaf dimension value combinations. The result is shown in Fig.~\ref{fig:time}. For the sake of analysis, the time in the figure is logarithmic. Obviously, all the methods will cost more inference time and exhibit the exponential trends to analyze the root causes when more dimensions are collected. In these methods, Adtributor is the most time-consuming method in all the situations which is followed by HotSpot. These two methods cost more than 500 seconds for $10^5$ leaf dimension value combinations. Although Squeeze and AutoRoot may be quick in less dimension value combinations, our CMMD can also provide competitive performance and achieve the best efficiency in the largest case. This is mainly due to the effect of the genetic algorithm on such combinatorial explosion problems to save the tedious traversal time in the tree-based search.

\section{Industrial Practice}
CMMD has been released as a private preview feature on an online service in Microsoft Azure which provides users to detect the anomaly and analyze root causes. Currently, it has been used by several customers from different backgrounds, including business marketing, AIOps(artificial intelligence for IT operations), and networking. 
In this section, we will analyze two real cases and share some lessons we learned from practice.


\subsection{Case 1: Business marketing} 

Conversion rate is one of the most commonly used and valuable KPIs to understand and effectively analyze the results of the marketing campaigns, which is derived by $\frac{\text{\#Conversions}}{\text{\#Views}}$. In this case, we monitor the Conversion Rate of a business system. There are two dimensions (Region and Channel) aggregated.

\begin{figure*}[h]
  \centering
  \includegraphics[width=0.75\textwidth]{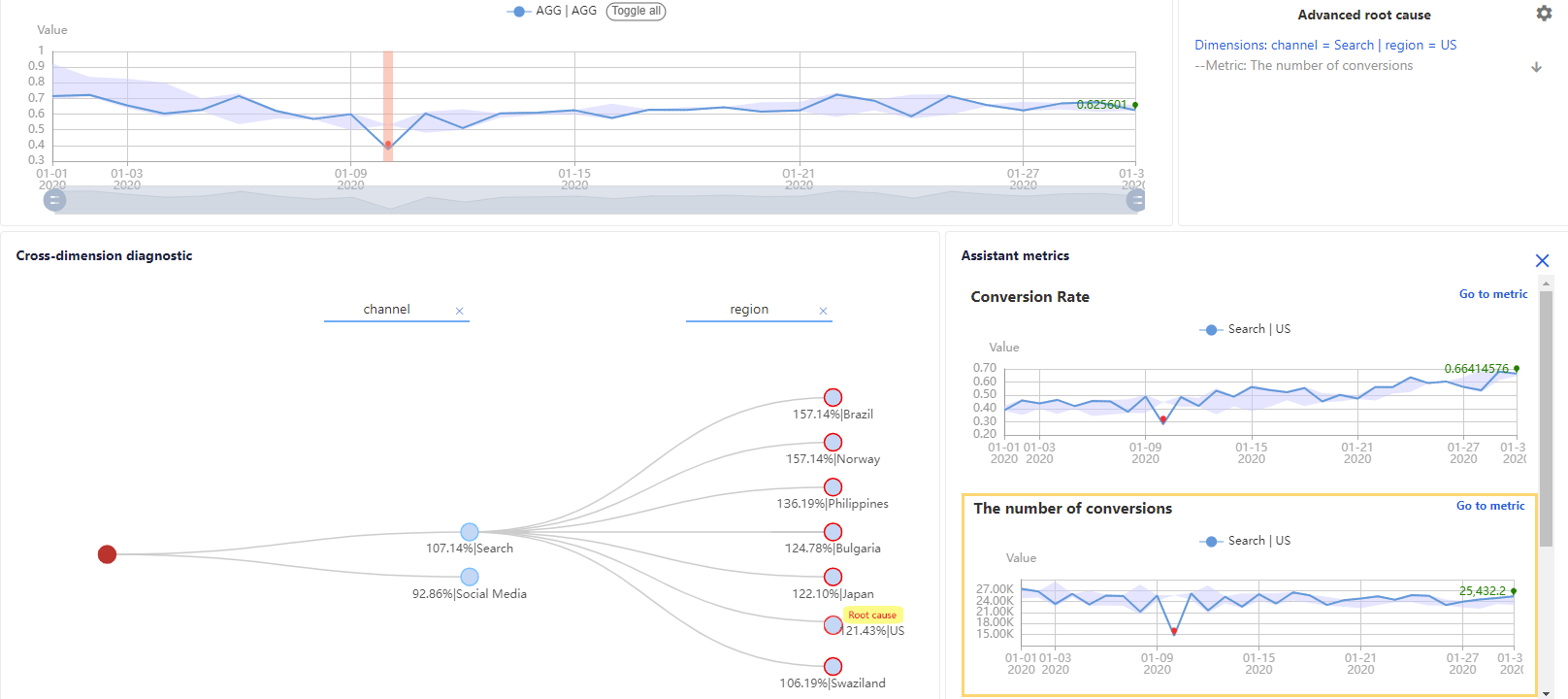}  
  \caption{Interface of our root cause service. 
  (x\% is equal to $\frac{\text{the value of node - the expected value of node}}{\text{the value of root node - the expected value of root node}}$. Previous methods found Brazil and Norway with their higher x\%. CMMD believes US is the root cause by the absolute value of \#Conversions.) }
  \label{fig:case_study}  
\end{figure*}

Fig.~\ref{fig:case_study} shows the interface of our service based on CMMD. At the top of Fig.~\ref{fig:case_study}, Conversion Rate of AGG|AGG was monitored with a 26.92\% decrease on January 10, 2020. 
The module of root cause analysis is triggered to find out the abnormal dimensions. Traditional methods solved this task only based on Conversion Rate data. Squeeze found two root causes: Search|Brazil with a 66.67\% decrease and Search|Norway with a 46.81\% decline both in Conversion Rate. 
HotSpot and Adtributor also regarded Search|Brazil as the root cause.
AutoRoot searched out an aggregated result Search|AGG with a 34.88\% drop.
However, Search|Brazil or Search|AGG are not the real root cause because: 
1) Although Search|Brazil changed much in Conversion Rate, its changes in \#Conversions have less impact (2.53\%) on \#Conversions of AGG|AGG. Thus, it also has a lower proportion to influence Conversion Rate of AGG|AGG;
2) 220 leaf dimension value combinations are involved in Search|AGG and only 3.2\% of them were abnormal at that timestamp. 

Different from previous works, CMMD believed that Search|US is the real root cause.
The 43.08\% decrease in \#Conversions of Search|US caused a 20.41\% drop in \#Conversions of AGG|AGG, and then led to a 26.92\% decrease in Conversion Rate of AGG|AGG.
Therefore, we show that Search|US is the root cause in the bottom left of Fig.~\ref{fig:case_study}. Besides, we also show the trends of corresponding metrics in the bottom right.
With CMMD, our service can point customers to the real root cause in one step. It was also confirmed by the data owners that the drop was caused by a conversion issue from one of the Search channels.



\subsection{Case 2: AIOps}

AIOps (Artificial Intelligence for IT Operations) is to enhance IT operations and guarantee the performance and reliability of services via AI techniques. CMMD also works well to monitor the status of services and help to restore the service performance when the anomaly occurs. In this case, we applied CMMD into an alerting service. We monitor the metric, Alert Open Rate, which is calculated by
$\frac{\text{\#Opened Alerts}}{\text{\#Alerts}}$. There is one dimension notification channel = \{Email, Azure DevOps, Microsoft Teams,...\}. 

A typical case is root cause analysis for abnormal Alert Open Rate of AGG with a 35.7\% drop (0.28 $\rightarrow$ 0.18). Other methods found DevOps as the root cause since DevOps has the largest deviation of Alert Open Rate. Although the change of Alert Open Rate of Email is only a half of DevOps, CMMD identified Email as a real abnormal channel because its deviation of \#Alerts contributes over 67.2\% to the anomaly compared with 2.3\% in DevOps. 
It also has been confirmed by our customers. At that timestamp, they sent many duplicate alerts in Email channel due to an updated version, which leads to the drop of Alert Open Rate.



\subsection{Lessons learned}
Although a scenario can contain a large volume of leaf dimension value combinations, not all the combinations have values at every timestamp. In other words, sparsity usually exists in the real industry. Therefore, it is unnecessary to monitor every leaf dimension value combination. The best practice is to aggregate the dimensions, monitor the aggregated metrics and analyze the root causes once the anomaly is detected.

In general, most of the KPIs that customers care about are not fundamental metrics, especially in the large-scale scenario. It is more important to monitor the derived metrics since these metrics are often related to profit or production safety. As can be seen in our previous analysis, considering only derived metrics is not enough for root cause analysis and the cross-metric algorithm can help find more accurate and fine-grained root causes.

\section{Conclusion}
In this paper, we proposed CMMD for cross-metric multiple dimensional root causes analysis. To our best knowledge, it is the first time to study the cross-metric problems which can accurately explain the anomalies occurred in derived metrics and find the root causes. Firstly, CMMD models the relationship between fundamental and derived metrics based on graph attention network. Based on the well-trained GAT, we adopt the genetic algorithm and design a novel fitness function to evaluate the probability of a set of dimension value combinations to be the root cause. We also use the attention mechanism in GAT for speed-up. Experiments suggest that CMMD can obtain better performance over other methods on average F1 scores and further explain the changes of monitored metrics. Through the case studies in production datasets, CMMD can locate the root causes more 
accurately and meet the requirements of industrial practice. 


\bibliographystyle{ACM-Reference-Format}
\bibliography{sample-base}

\clearpage
\appendix

\section{Supplementary materials}

\subsection{Pseudocode}
\begin{algorithm}
    \caption{GA-based root cause localization}
    \label{algo:GA}
    \begin{algorithmic}[1]
        \REQUIRE Candidate dimension combinations, Graph attention model $\mathcal{G}$, population size $N_p$, iterations $N$, crossover rate $t_c$, mutation rate $t_f$
        \ENSURE root cause set 
        \STATE Initialize a population $\boldsymbol{S}$ with $N_p$ binary chromosomes;
        \FOR{$i=1 \to N$}
            \FOR{$j=1 \to N_p$}
                \STATE Compute the fitness score for each $s_j$ in $\boldsymbol{S}$ by Eq.~\ref{eq:fs}
            \ENDFOR
            \STATE Record the optimal $s^*$ with the smallest fitness score;
            \STATE Select new population $\boldsymbol{S}^{\prime}$ based on roulette where the probability is consistent with fitness score; 
            \STATE Cross random part of each two chromosomes in $\boldsymbol{S}^{\prime}$ with probability $t_c$;
            \STATE Mutate each chromosome in $\boldsymbol{S}^{\prime}$ with probability $t_f$;
            \STATE Replace the population $\boldsymbol{S}$ with $\boldsymbol{S}^{\prime}$ 
        \ENDFOR
        \STATE Generate the root cause set $\mathcal{S}$ based on $s^*$
    \end{algorithmic}
\end{algorithm}

\begin{algorithm}
    \caption{Root cause backtrack}
    \label{algo:backtrack}
    \begin{algorithmic}[1]
        \REQUIRE Root cause set $\mathcal{S}$, number of layers $L$ in dimension tree $\mathbb{T}$, backtrack threshold $t_{\gamma}$
        \ENSURE A succinct set of root causes $\mathcal{S}^*$   
        \STATE $\mathcal{S}^* = \mathcal{S}$
        \FOR{$l=L-1$ to $1$}
            \STATE Search the parent node set $PS$ of $\mathcal{S}^*$ in layer $l$ of $\mathbb{T}$;
            \FOR{$p \in PS$}
                \STATE Search leaf nodes belonging to $p$ as $a$;
                \STATE Search the leaf nodes belonging to $p$ and also in $\mathcal{S}^*$ as $b$;
                \IF{$\frac{\text{Count}(b)}{\text{Count}(a)} \geq t_{\gamma}$}
                    \STATE $\mathcal{S}^*.add(p)$
                    \STATE $\mathcal{S}^*.remove(b)$
                \ENDIF
            \ENDFOR
            \IF{None of $PS$ is added to $\mathcal{S}^*$}
                \STATE Break
            \ENDIF
        \ENDFOR
    \RETURN $\mathcal{S}^*$  
    \end{algorithmic}
\end{algorithm}

\subsection{Implementation details}
All the experiments are implemented by Python 3.8.8 in a Laptop with Intel(R) Core(TM) i7-8665U CPU @2.11GHz and 16G RAM. 
\subsubsection*{Adtribtor}
Adtributor locates the root causes by dimension. Values in each dimension are sorted by Jansen-Shannon divergence and the stop searching criteria within dimension values and dimensions are both controlled by thresholds of cumulative explanatory power. The threshold for explanatory power to stop searching in a dimension is specified as $T_{EEP} = 0.3$ and threshold for cumulative explanatory power of dimension values is specified as $T_{EP} = 0.8$.
\subsubsection*{HotSpot}
HotSpot proposes the \textbf{potential score} to evaluate the dimension value combination and is used to sort the candidate dimension value combinations. In HotSpot, Monte Carlo Tree search is utilized to locate the root causes in an efficient way. Refer to the settings in \cite{sun2018hotspot}, the maximum iteration $M$ and stop condition $PT$ which control the  \textbf{potential score} of dimension value combinations are specified as $M=25$ and $PT=0.99$.
\subsubsection*{Squeeze}
Squeeze analyzes the root cause in two procedures. Firstly, deviations between real and expected values of metrics in dimension value combinations are filtered and clustered into several clusters. The threshold in Squeeze to filter abnormal dimension values is automatically determined by \textit{knee-point} method and the \textit{bins} in clustering is set as 30. In the search process, $\theta = 0.9$ is set to control \textbf{generalized potential score} of candidate dimension values in the top-bottom strategy. Finally, the constant $C$ for succinctness is calculated based on the original paper. 
\subsubsection*{AutoRoot}
AutoRoot also adopts two procedures similar to Squeeze. Differently, kernel density estimation is used in clustering to determine the number of clusters based on the distribution of deviations of dimension values. The $\text{bandwith} = 0.2$ is set in the kernel function to estimate the distribution. Secondly, $\theta = 0.9$ is set as the threshold of \textbf{new potential score}.
\subsubsection*{CMMD}
The proposed CMMD mainly consists of two modules, i.e., relationship modeling and root cause localization. The hyper-parameters used in these two modules are summarized in Table~\ref{tab:hyper}. During GAT training stage, the batch size is equal to the number of non-leaf nodes in the dimension tree. In root cause localization, the number of population $N_p$ and the number of iteration $N$ are depended on the size of data. $N_p$ can be set from 10 to 50 and $N$ can be set from 5 to 10 in the experiments.
The length of chromosomes in GA is equal to the number of dimension value combinations after filtering. $t_\delta$ determines the abnormal degree of nodes in candidate set and we search $t_\delta$ from {0.1,0.2} for different datasets in this work. 

\begin{table} \small
  \caption{Hyper-parameters in CMMD}
  \label{tab:hyper}
  \begin{tabular}{c|c|c}
    \toprule
    Component &  Hyper-parameter & Value\\
    \midrule
        & Number of epochs & 1000\\
        & Embedding dimension & 8 \\
    GAT & Number of attention heads & 8\\
        & Learning rate & 5e-4\\
        & Optimizer & Adam\\
        & Early stop patience & 50\\
    \midrule
       & Cross rate $t_c$ & 0.5\\
    GA & Mutation rate $t_f$ & 0.1\\
       & $\beta$ in Eq.~\ref{eq:fs} & 1.0\\
    \midrule   
    Others  & Backtrack threshold $t_{\gamma}$ & 0.6\\
       & Filtering threshold  $t_{\delta}$ & Searched from {0.1, 0.2} \\
  \bottomrule
\end{tabular}
\end{table}

\subsection{Details in evaluation}
In the experiments, evaluation metrics are calculated between the ground truth and the output of these methods. In the three types of datasets, the ground truth of $\mathcal{D}_2$ is provided in \cite{li2019generic}. In other two datasets, we assume that we know the calculation and aggregation function. Thus, the ground truth is a set of dimension value combinations that has the highest recovery ratio based on the known calculation and aggregation function. In detail, we firstly replace the real values with expected values of fundamental metrics in the candidate set of dimension value combinations and then calculate the recovery values of the monitored metric by aggregation function. Further, the recovery ratio with and without replacement is also calculated. If the recovery value is closer to the expected value, the recovery ratio is higher which means that these dimension value combinations can explain the anomaly in the monitored metric better. Given the fact that every dimension value combination can contribute to the abnormal monitored metric more or less, we choose the dimension value combination(s) whose cumulative recovery ratio exceeds a certain threshold (80\% in this paper) as the ground truth.

\begin{figure*}[t!]
  \centering
  \subfigure[Precision]{
  \begin{minipage}[b]{0.3\textwidth}
  \label{fig:beta1}
  \includegraphics[width=\linewidth]{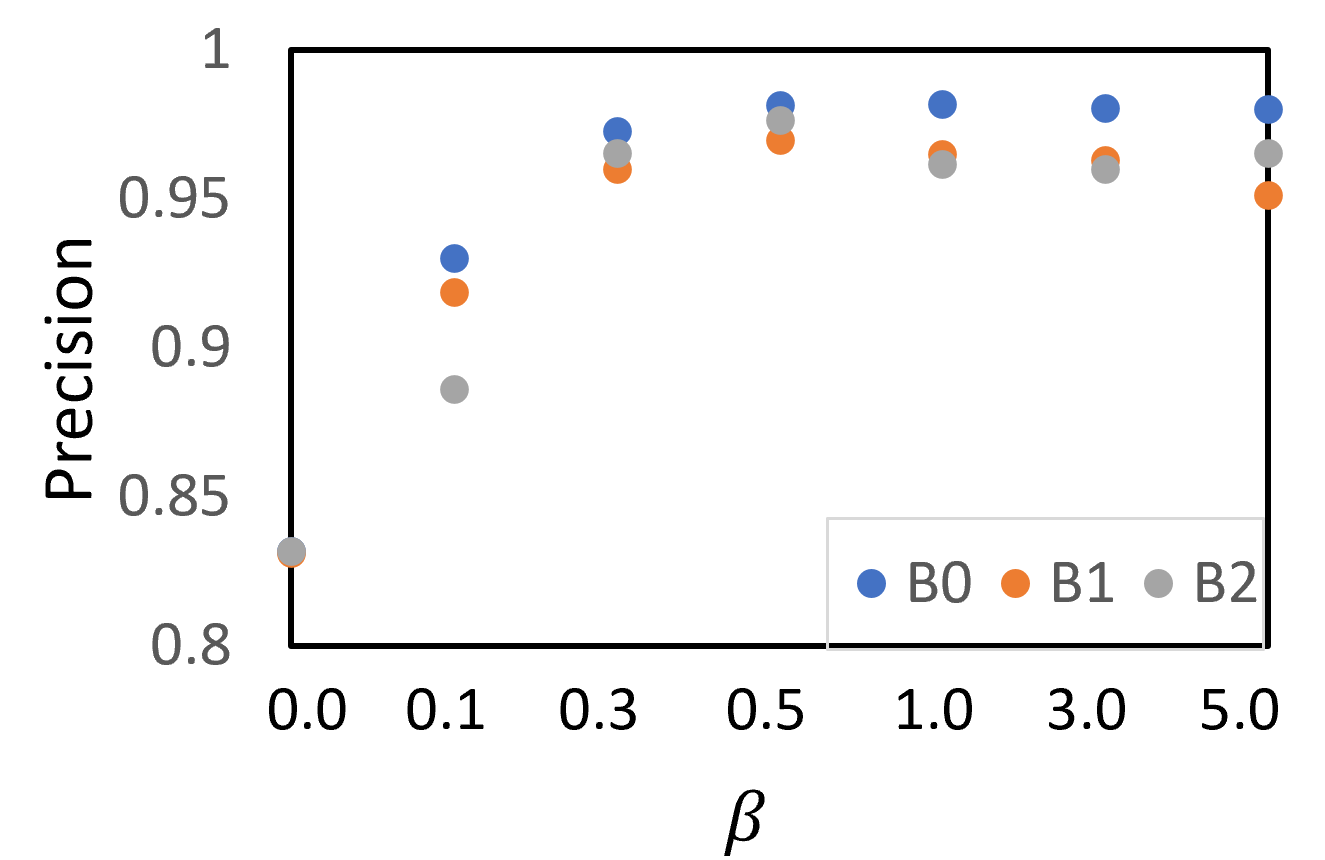}
  \end{minipage}
  }
  \subfigure[Recall]{
  \begin{minipage}[b]{0.3\textwidth}
  \label{fig:beta2}
  \includegraphics[width=\linewidth]{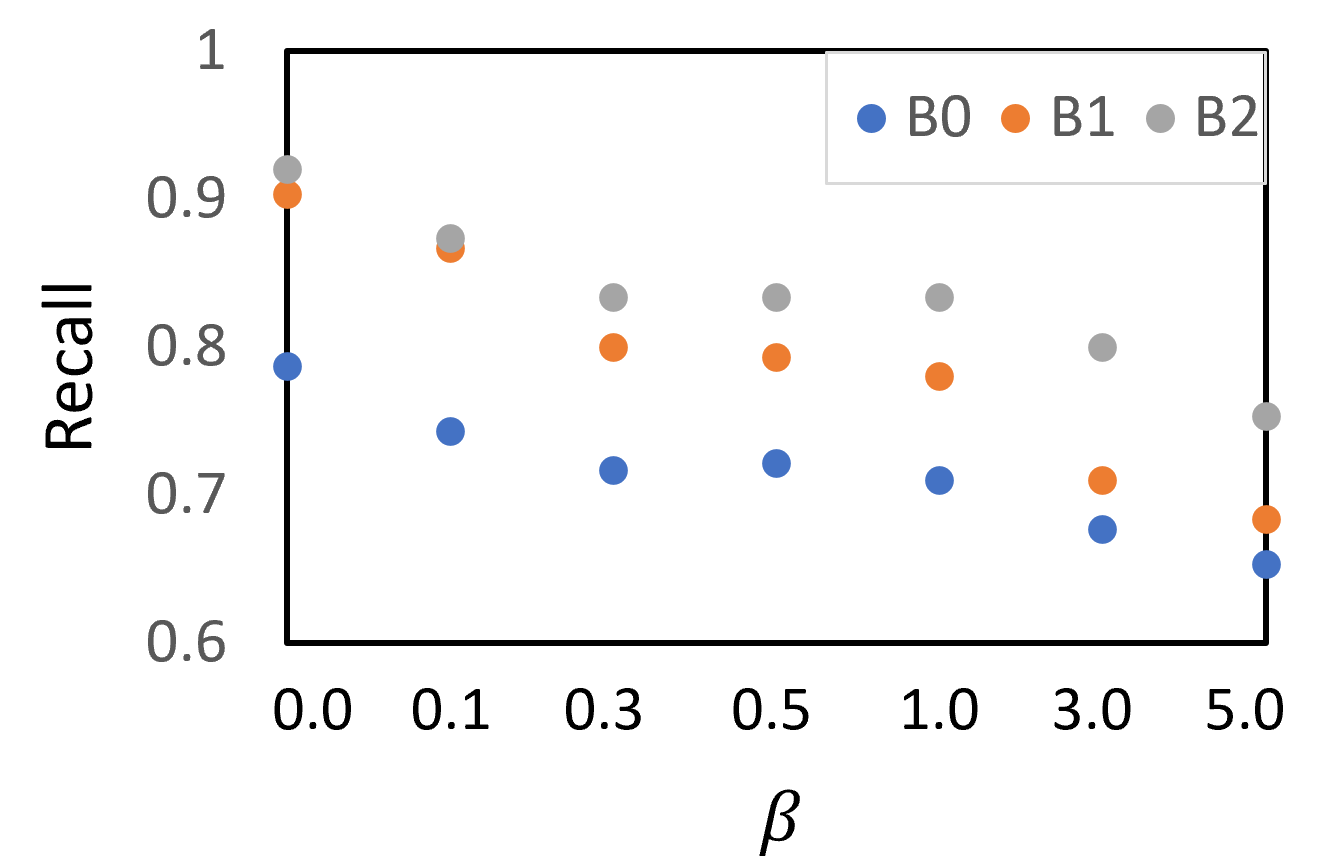}
  \end{minipage}
  }
  \subfigure[F1 score]{
  \begin{minipage}[b]{0.3\textwidth}
  \label{fig:beta3}
  \includegraphics[width=\linewidth]{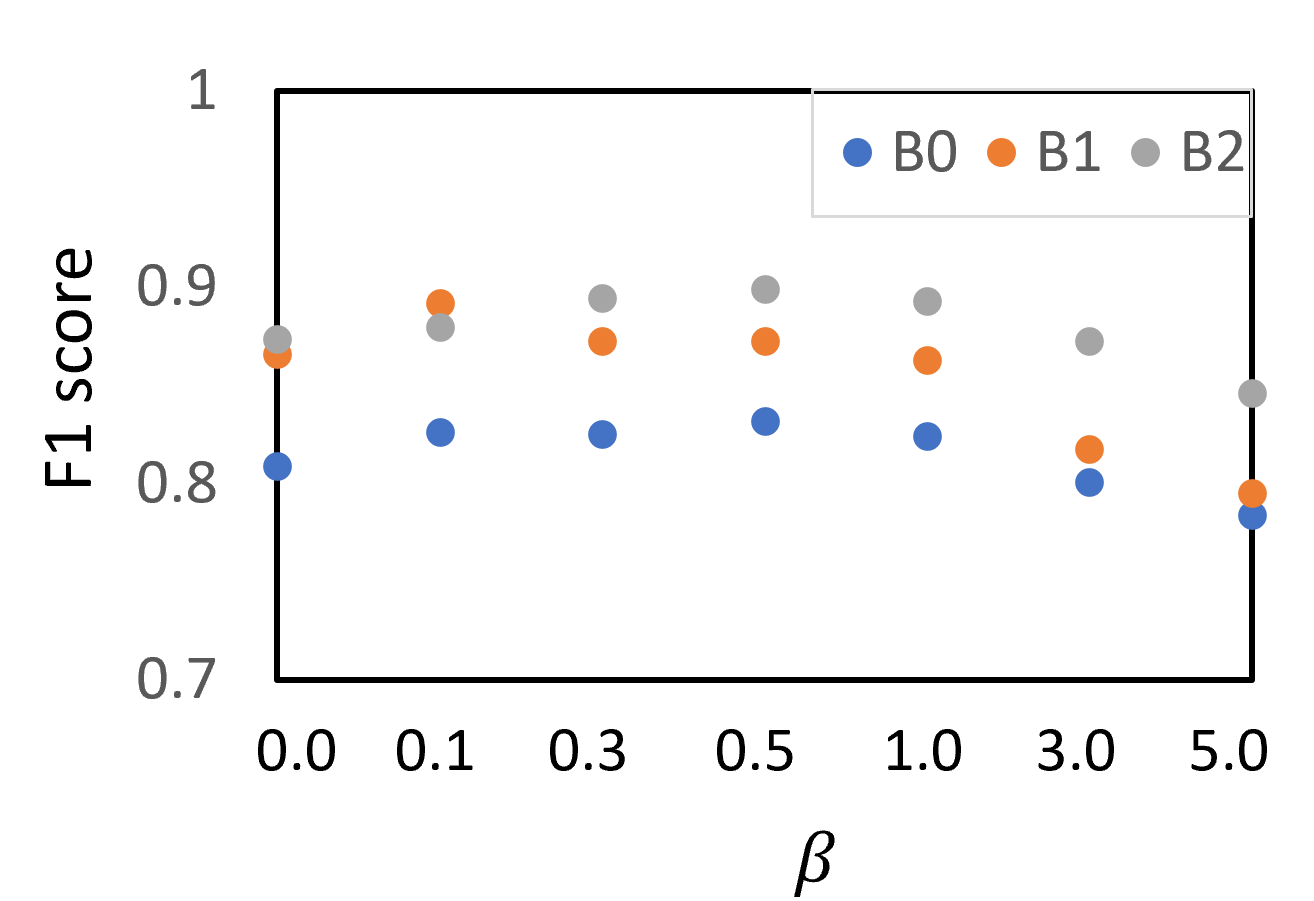}
  \end{minipage}
  }
  \caption{Performance over different $\beta$}
  \label{fig:beta}  
\end{figure*}

\subsection{Synthetic datasets}
The synthetic datasets $\mathcal{D}_3$ in the experiment are generated in the following steps:
\begin{itemize}
    \item [(1)] 
    Determine the number of dimensions and values in each dimension to form the dimension tree $\mathbb{T}$
    \item[(2)] 
    Determine the metric relationship as follows. In $\mathcal{D}_3$, we have two fundamental metrics $b$ and $c$, and two derived metrics $a=g(c)$ and $d = f(a,b)$.
    \item[(3)] 
    Generate the data of $b$ and $c$ in each dimension value combination. Here we firstly use the uniform distribution to generate the value of $b$ and $c$ at specific timestamps. Then we randomly choose an operator from $g=\left\{c, \sin(c), e^c, c^2, \sqrt{c} \right\}$ to transform $c$ into $a$. Finally, we set function $f$ as five derived functions: 1) $\frac{a}{b}$, 2) $a \times b$, 3) $\frac{\log a}{\log b}$, 4) $a \times e^b$ and 5) $\frac{\log (a+1)}{\log (b+1)}$.
    \item[(4)] 
    Aggregate the fundamental metrics in the dimension tree $\mathbb{T}$ and then calculate the corresponding derived metrics. 
    \item[(5)] 
    Inject the anomalies randomly in the fundamental metrics of selected dimension value combinations and then calculate the abnormal metrics of corresponding dimension values.
    \item[(6)] 
    Calculate the expected values of each metric by the auto regressive models and detect the anomalies by the 3$\sigma$ principal.
\end{itemize}

\subsection{Parameter configuration}
The parameter configurations are important for the existing root cause analysis approaches as well as the proposed CMMD. Here we consider the important trade-off parameter $\beta$ in Eq.~\ref{eq:fs} for succinctness of root causes. To conduct experiments under different values of $\beta$, we select subdatasets B0,B1,B2 in~\cite{li2019generic}. The results under different parameter values are shown in Fig.~\ref{fig:beta}. In general, with the increase of $\beta$, the strategy focuses more and more on the succinctness rather than the scores of root causes which will lead to the degradation of performance. Specifically, Precision and Recall show the opposite trend and we can find a better F1 score when $\beta \in [0.1, 1.0]$ and CMMD is somehow robust to this hyperparameter. Therefore it is not difficult to configure this parameter in practical.

\end{document}